\documentclass{article}

\usepackage{arxiv}

\usepackage[utf8]{inputenc} 
\usepackage[T1]{fontenc}    
\usepackage{hyperref}       
\usepackage{url}            
\usepackage{booktabs}       
\usepackage{amsfonts}       
\usepackage{nicefrac}       
\usepackage{microtype}      
\usepackage{lipsum}		
\usepackage{doi}
\usepackage{graphicx}
\usepackage{epstopdf}
\usepackage{subcaption}
\usepackage{framed,multirow}
\usepackage[english]{babel}

\title{Face Anti-Spoofing from the Perspective of Data Sampling}

\date{} 					

\author{ \href{https://orcid.org/0000-0001-7191-0245}{\includegraphics[scale=0.06]{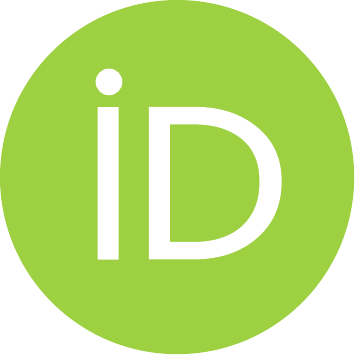}\hspace{1mm}Usman Muhammad and Mourad Oussalah}\thanks{Corresponding author: Usman Muhammad} \\
	Center for Machine Vision and Signal Analysis\\
	University of Oulu\\
	\texttt{Muhammad.usman@oulu.fi} \\
}

\renewcommand{\shorttitle}{\textit{arXiv} Template}

\hypersetup{
pdftitle={Face Anti-Spoofing from the Perspective of Data Sampling},
pdfsubject={q-bio.NC, q-bio.QM},
pdfauthor={Usman Muhammad, Mourad Oussalah},
pdfkeywords={First keyword, Second keyword, More},
}

\begin{document}
\maketitle

\begin{abstract}	
Without deploying face anti-spoofing countermeasures, face recognition systems can be spoofed by presenting a printed photo, a video, or a silicon mask of a genuine user. Thus, face presentation attack detection (PAD) plays a vital role in providing secure facial access to digital devices. Most existing video-based PAD countermeasures lack the ability to cope with long-range temporal variations in videos. Moreover, the key-frame \textit{sampling} prior to the feature extraction step has not been widely studied in the face anti-spoofing domain. To mitigate these issues, this paper provides a data sampling approach by proposing a video processing scheme that models the long-range temporal variations based on Gaussian Weighting Function. Specifically, the proposed scheme encodes the consecutive $t$ frames of video sequences into a single RGB image based on a Gaussian-weighted summation of the $t$ frames. Using simply the data sampling scheme alone, we demonstrate that state-of-the-art performance can be achieved without any bells and whistles in both intra-database and inter-database testing scenarios for the three public benchmark datasets; namely, Replay-Attack, MSU-MFSD, and CASIA-FASD. In particular, the proposed scheme provides a much lower error (from 15.2\% to 6.7\% on CASIA-FASD and 5.9\% to 4.9\% on Replay-Attack) compared to baselines in cross-database scenarios.
\end{abstract}


\section{Introduction}

During the past decade, we have witnessed great advancements in face anti-spoofing methods thanks to the emergence of deep learning models. However, with the growth of facial recognition technology, face identity threats are evenly increasing at the same rate and the problem is still unsolved due to the difficulty in the design of discriminative features. Therefore, how to effectively detect spoofing attacks remains a critical problem for both practitioners and the research community. Existing learning-based approaches can be classified based on static and dynamic information. Compared to static or image-based face presentation attack detection (PAD) \cite{1}, video-based face anti-spoofing is more challenging because deep learning methods based on a 2D convolutional neural network (CNN) ignore the temporal dimension of the video and process each frame independently.

Although spatiotemporal feature learning based on optical flow \cite{2}, 3D CNN \cite{3} or recurrent neural network (RNN) \cite{4} have demonstrated their performance in the face anti-spoofing domain, there is a fundamental question that still needs to be addressed in the field -What are alternative approaches to improve the spatiotemporal representations for face anti-spoofing? Wang {\textit{et al}} \cite{5} argued that temporal depth difference between live and spoof faces along with a contrastive depth loss can make impressive progress in improving the PAD performance. Zitong {\textit{et al}} \cite{6} proposed a method based on central difference convolution (CDC) via aggregating both intensity and gradient information. Recently, self-supervised learning \cite{7}, meta-teacher learning, or single-shot face anti-spoofing \cite{8} were introduced for improving PAD performance. However, there is no clear winner among spatiotemporal feature learning based-methods in terms of accuracy.

We argue that most of the existing video-based countermeasures employ a fixed frame selection and overlook other important factors such as the importance of data understanding. Moreover, within a given video, not all the frames are of equal importance for PAD detection, and action-related frames may occur sparsely in a few frames. Thus, the key-frame \textit{sampling} \cite{9} is crucial before the feature extraction step and it remains an unsolved problem for video-based PAD. Although our previous works \cite{4,7} demonstrate that the spoofed videos are consistent with motion cues associated with the artifacts, i.e., hand trembling or material reflection, such methods require complex feature engineering skills to generate the spatiotemporal representations. 
 \begin{figure}[h]
\centering
    \includegraphics[width=12.5cm]{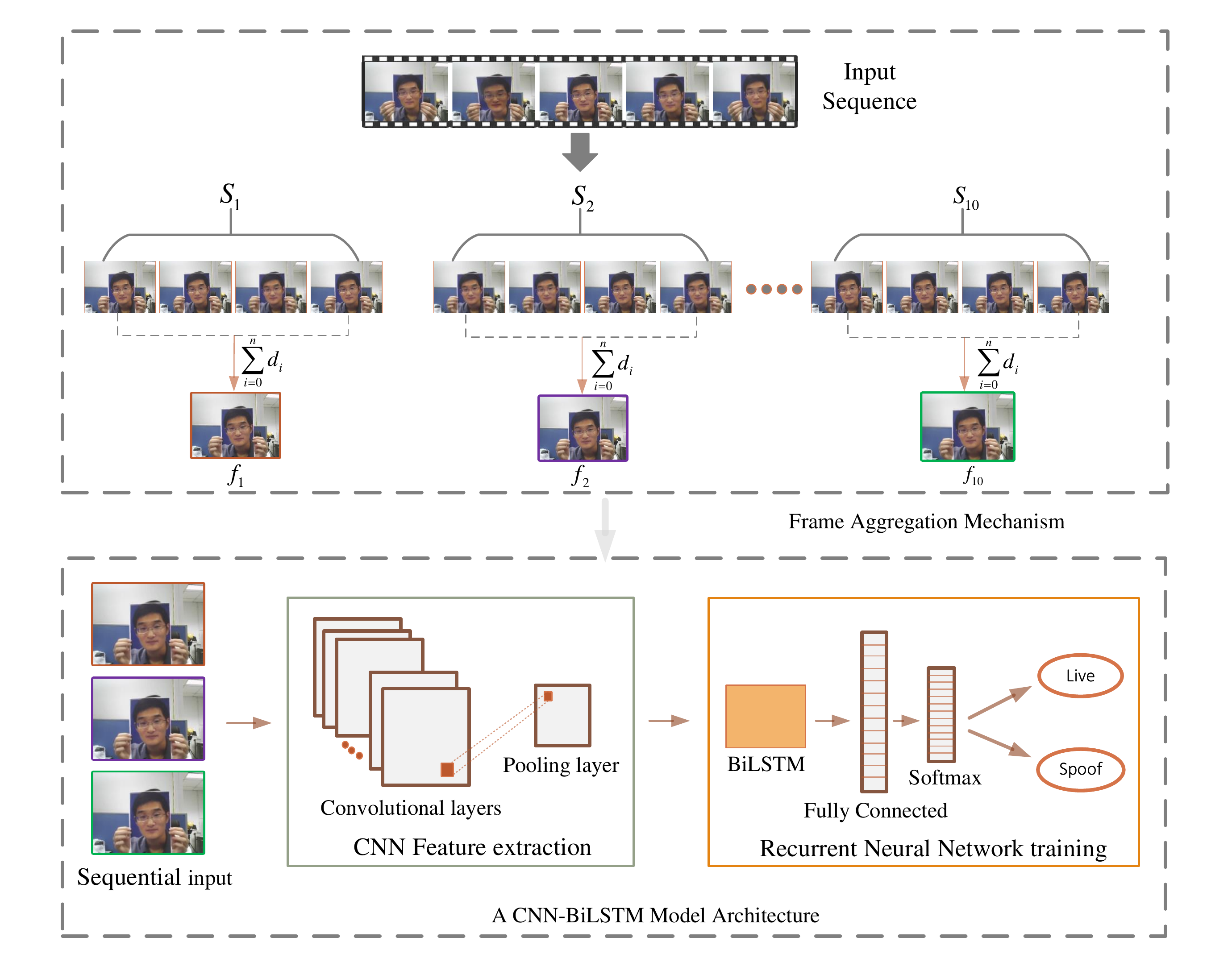}
    \caption{An illustration of the proposed framework.}
\end{figure}
As a result, we need to develop a simple strategy of selecting relevant frames according to their temporal variations. Thus, using simply the Gaussian Weighting Function \cite{10}, we propose a simple data sampling mechanism, which aims to accumulate information from video sequences into a single RGB image and generate the most discriminative frames. Specifically, we demonstrate that appropriate data sampling provides an alternative solution to enhance the existing deep learning-based face anti-spoofing. Our main objective is to provide the importance of data understanding, especially, i) how differently does CNN or RNN behave concerning spatial-temporal modeling of video data? and ii) conduct comprehensive experiments and analysis to demonstrate that our approach is extensible, which can be plugged into different deep learning-based face PAD models. In summary, the overall contributions of our work include:
\begin{enumerate}
\item We present a data-driven approach to capture the appearance and dynamics of video into a single RGB image. 

\item Our analysis shows that the proposed temporal modeling can amplify important clues, e.g., hand movements, and surface edges, to improve the detection accuracy.

\item We provide an interpretation of the decisions made by the employed model. The model revealed that the motion cues are the most important factors for distinguishing whether an input image is spoofed or not. 

\item Experiments on three benchmark datasets, consisting of CASIA-MFSD, Replay-Attack, and MSU-MFSD databases, show that our proposed method is significantly superior to the state-of-the-art generalization methods used now.
\end{enumerate}

\section{Proposed Method}
The proposed data sampling is based on the assumption that adjacent frames in videos complement the temporal motion which is an important and universal signal. This also motivates us to adaptively select frames from videos. To achieve this, Gaussian Weighting Function (GWF) is used to accumulate the video sequences into a fewer number of frames. Suppose that $\{S_n\}n \in N$ is an exhaustive non-overlapping sequence (pertaining to the full length of the video), which is given by

\begin{equation}\label{eq1}
S_{n} = \{S_{1},S_{2}, \dots, S_{k},\dots\},
\end{equation}
Where $\{S_k\}$ represents the $k^{th}$ sub-sequence of $\{S_n\}$ and $k<n$. Specifically, Gaussian Weighting Function G, for a sub-sequence $\{S_k\}$, is denoted as follows:
\begin{equation} \label{eq2}
G(S_{k}, M) =  \sum_{q=1}^{Z} S_{k_q} \ast \frac{M_q} {\sum_{q=1}^{Z}{M_{q}}}
\end{equation}

The function $G$ takes a sub-sequence $\{S_k\}$, and Gaussian weight vector $M$ as input, and accumulates the representation into a single RGB frame. In particular, $M_{q}$ depicts the $q^{th}$ element of the Gaussian weight vector $M$. $Z$ represents the size of the Gaussian weight vector. For instance, if the size of the Gaussian weight vector $Z$ is 30 and the sub-sequence $\{S_k\}$ has thirty frames of the video. Then, the vector $M$ is given by $M$ = $[1, 2, 3,...,30]$. A single RGB image can be acquired based on the weighted summation of the thirty frames associated with the sub-sequence $\{S_k\}$ as illustrated in equation (2). Thus, following equation (2), different frame sizes can be selected to be accumulated into a single RGB image based on the GWF. Likewise, the same procedure is applied for the subsequent thirty frames belonging to the next sub-sequence and so on. The main steps are illustrated in Fig.1. Intuitively, this data sampling method has at least three main advantages. First, the encoded Spatio-temporal information can be fed to any CNN architecture for a still image, where “still” captures long-range dynamic variations in videos. Second, the encoded images decrease the number of frames per video to be processed and force the learning-based classifiers to focus on discriminative clips. Third, our method accumulates raw video frames directly in comparison to the previous data preprocessing methods that use complex processing such as estimation of global motion \cite{7} or optical flow \cite{2}.
\begin{table}[h]
\begin{center}
\caption{Ablation study using cross-database evaluation.}
\begin{tabular}{c |c| c}
  \hline
    \multirow{2}{*}{Method} & Train CASIA-FASD  & Train Replay-Attack \\
      & Test Replay-Attack  & Test CASIA-FASD \\
  \hline
                   Sub-sequence size (30)  & 3.1 &  15.3\\
                   Sub-sequence size (40)  & 4.9 &  6.7 \\
                   Sub-sequence size (50)  & 3.8 &  12.4 \\
                   Sub-sequence size (60)  & 4.6 &  10.1 \\
  \hline
 \end{tabular}
\end{center}
\end{table}

\section{Implementation details}
We conduct experiments on three major databases: CASIA Face Anti-Spoofing \cite{11}, Idiap Replay-Attack \cite{12}, and MSU Mobile Face Spoofing \cite{13}. In order to determine the performance, we report Half Total Error Rate (HTER), and Equal Error Rate (EER) by following the existing testing protocols used in the current works \cite{27,5}. A pre-trained CNN (ResNet-101 \cite{14}) is utilized to extract off-the-shelf CNN features using the pooling layer without data augmentation. All the images are resized to $224 \times 224$. The extracted features are then used as input to train the Bidirectional Long ShortTerm Memory Networks (BiLSTM) \cite{15} for final detection. To train the BiLSTM, the Adam optimizer is utilized by fixing a learning rate of $0.001$, mini-batch size $32$, validation frequency $30$, and with $100$ hidden layer dimension for all the intra-database protocols. An early stopping function \cite{16} is utilized to reduce the risk of over-fitting since we do not set the fixed number of epochs. For the cross-database or inter-database scenario, the same experimental settings are used except the learning rate which was increased up to $0.0001$ for the BiLSTM. Moreover, {\textit{He}} initializer \cite{17} is used for initializing recurrent weights that perform the best for all our experiments.
\begin{figure}
\centering
\begin{subfigure}[b]{0.45\textwidth}
   \includegraphics[width=1\linewidth]{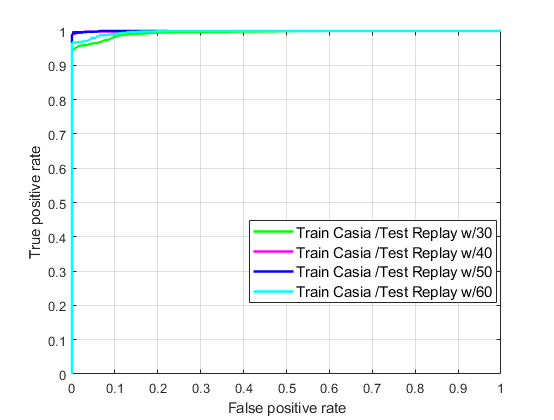}
   \caption{}
   \label{fig:Ng1} 
\end{subfigure}
\begin{subfigure}[b]{0.45\textwidth}
  \includegraphics[width=1\linewidth]{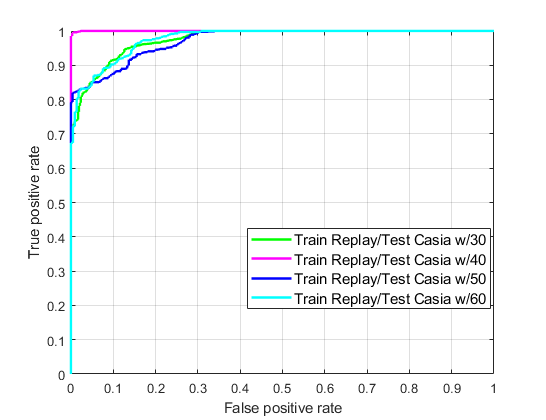}
  \caption{}
   \label{fig:Ng2}
\end{subfigure}
\caption{The Receiver Operating Characteristics (ROC) curves of the (a) Replay-Attack and (b) CASIA databases, where the true positive rate depicts the y-axis and false positive rate displays the x-axis.}
\end{figure}
\section{Ablation study}
To investigate the influence of temporal variations in the videos, we thoroughly validate the performance of each sub-sequence size by performing an ablation study. The numerical results are reported in Table 1. The result shows that even capturing the small temporal variations leads to a competitive performance by using our proposed data sampling. It is worth mentioning that the model is trained on the training dataset of CASIA and uses the CASIA testing set as a validation set. Based on the EER of the CASIA testing set, we report HTER on a completely unseen Replay-Attack dataset. Similarly, the same experimental protocols are repeated for the Replay-Attack dataset. We note that the proposed data sampling provides consistent performance even with a sub-sequence size of 30. When the temporal length increases to 40, we achieved the best performance on the CASIA dataset and very competitive performance for the Replay-Attack dataset. Thus, we set the same sub-sequence size (40) for the intra-database evaluation. For further analysis, ROC curves are presented in Fig.2. The true positive rate (TPR) and false positive rate (FPR) allow a learning curve in the upper left corner in case the model provides a higher sensitivity rate. Through Fig.2, we can see that the model classifies robustly in both databases.

\begin{table}[h]
\centering
    \caption{The results of intra-database evaluation.}
\begin{tabular}{c|c|c|c|c}\hline & \multicolumn{2}{c|}{REPLAY-ATTACK} & \multicolumn{1}{c|}{CASIA}  & \multicolumn{1}{c}{MSU} \\  \hline
{Methods} & {EER$(\%)$}  & {HTER $(\%)$} & EER$(\%)$ & EER$(\%)$ \\ \hline
 IDA \cite{13}   & - & - & 12.97 & 8.58 \\ 
 Color LBP \cite{18}  & 37.9  & 21.0 & 35.4 & 44.8 \\ 
  Patch-CNN \cite{19}   & 0.72 & -  & 4.44 & - \\ 
  Hybrid CNN \cite{20}   & - & -  & 0.02 & 0.04 \\ 
  SLRNN \cite{21}   & - & -  & 0.01 & 0.02 \\ 
  SPMT + SSD \cite{22}   & 0.04 & 0.06 & 0.04 & - \\ 
   GFA-CNN \cite{23}   & 0.30  & -  & 8.3 & 7.5 \\ 
    S-CNN \cite{24}  & 0.28  & -  & 0.53 & 0.18 \\ 
 \hline
ResNet-BiLSTM w/DS & \textbf{0.00}  & \textbf{0.15} & \textbf{0.00} & \textbf{0.00} \\ \hline
            \end{tabular}
\end{table}

\begin{table}[h]
\begin{center}
\caption{The results of cross-database testing.}
\begin{tabular}{c |c| c}
  \hline
    \multirow{2}{*}{Method} & Train CASIA-FASD  & Train Replay-Attack \\
      & Test Replay-Attack  & Test CASIA-FASD \\
  \hline
    Color-LBP \cite{18}  & 30.3 &  37.7 \\
    Auxiliary \cite{25}  & 27.6  &  28.4 \\
      FaceDs \cite{26}  & 28.5 & 41.1 \\
       STASN \cite{27}   & 31.5 &  30.9 \\
     DSGTD \cite{5} & 17.0 & 22.8 \\
        CDCN \cite{6}  & 6.5 &  29.8\\
          SSL learning \cite{7} & 5.9 & 15.2 \\
          \hline
          VGG19-BiLSTM w/o DS  & 31.3 & 42.8 \\
           VGG19-BiLSTM w/DS  & 13.3 & \textbf{11.4} \\
           ResNet-BiLSTM w/o DS  & 27.4 & 33.2 \\
           ResNet-BiLSTM w/DS & \textbf{4.9} & \textbf{6.7} \\
  \hline
 \end{tabular}
\end{center}
\end{table}

\section{Comparison against the state-of-the-art methods}
We compare the performance against the state-of-the-art methods in the intra-database scenario and report the results in Table 2. As it can be seen from Table 2, the model provides state-of-the-art results for the CASIA, REPLAY, and MSU datasets. Specifically, our method achieves $0.00$, EER for the Replay-Attack, CASIA, and MSU datasets, respectively. When we evaluated the most challenging scenario, i.e., cross-dataset in terms of HTER, a significant difference can be seen especially for the CASIA dataset in Table 3. It improves the performance by up to 8$\%$ more than the previous state-of-the-art method \cite{7}. This is a remarkable improvement for the cross-database scenario. Furthermore, we also report the results of the ResNet-BiLSTM model without our proposed data sampling. One can see that the performance is improved up to 20 $\%$ on both datasets. To demonstrate that the proposed data sampling can enhance the performance of other deep learning models, we also report the performance of VGG19 architecture \cite{28} using a sub-sequence size (40). The results confirm the effectiveness of our approach. Based on these experimental results, we demonstrate that sufficient data understanding is important in PAD detection. In addition, we provide heat map visualizations based on occlusion sensitivity maps \cite{29} and local interpretable model-agnostic explanations (LIME) \cite{30} to understand what patterns contribute to making a classification decision. The first column in Fig.3 (a) shows the images generated by the proposed data sampling scheme. The heat map in Fig.3 (b) based on occlusion sensitivity maps provides vital information for the model prediction. One can observe that the proposed frame accumulation contributes positively to distinguishing real and spoof faces. Specifically, images of photo and replay-attack in the first and second row show that hand movement provides salient information. For further evaluation, LIME interpretation in Fig.3 (c) is provided to demonstrate that the proposed data sampling helps the models to focus on the paper's texture and hand rotation cues. The masked images in Fig.3 (d) represent the most important superpixels to view the model's decision in a human-understandable way.

\begin{figure}[h]
\centering
    \includegraphics[width=9.8cm]{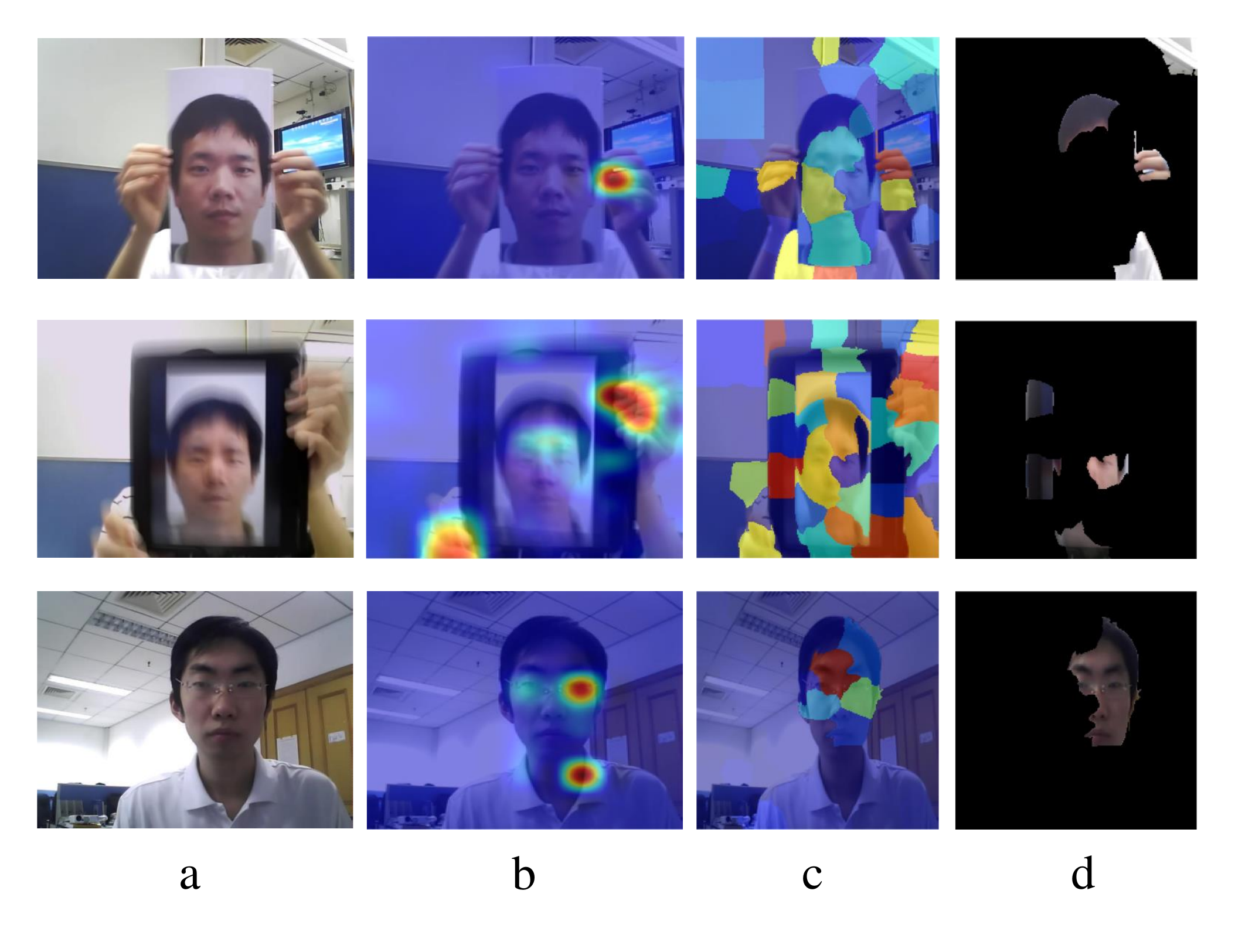}
    \caption{Image explanation using occlusion sensitivity maps and LIME corresponding to a print attack (first row), video-replay attack (second row) and real face (third row).}
\end{figure}

\section{Conclusions}
In this paper, we addressed the face PAD issue by proposing a simple data sampling strategy that requires a Gaussian weighting function to aggregate the video sub-sequences into a single image. Since the motion-based cues are naturally presented in the format of video streams, we explored the dynamic variations in video sub-sequences to design an effective frame selection scheme. Extensive experiments on three datasets demonstrate that our proposed method is robust in both intra-database and inter-database testing scenarios. Our future work includes utilizing video classification methods to represent videos in a more discriminative form.




\begin{thebibliography}{}


\bibitem{1}
Nong, X., Zeng, Y.,  Hu, H. (2021). Face anti‐spoofing with refined triplet loss and multi‐level attention constraint network. Electronics Letters, 57(24), 912-914.

\bibitem{2}
Li, L., Xia, Z., Wu, J., Yang, L., Han, H. (2022). Face presentation attack detection based on optical flow and texture analysis. \textit{Journal of King Saud University-Computer and Information Sciences}, \textbf{34(4)}, 1455-1467.

\bibitem{3}
Gan, J., Li, S., Zhai, Y., Liu, C. (2017, March). 3d convolutional neural network based on face anti-spoofing. \textit{In 2017 2nd international conference on multimedia and image processing (ICMIP)} (pp. 1-5). IEEE.

\bibitem{4}
Muhammad, U., Zhang, J., Liu, L.,  Oussalah, M. (2022). An Adaptive Spatio-temporal Global Sampling for Presentation Attack Detection. \textit{IEEE Transactions on Circuits and Systems II: Express Briefs}.

\bibitem{5}
Wang, Z., Yu, Z., Zhao, C., Zhu, X., Qin, Y., Zhou, Q., Lei, Z. (2020). Deep spatial gradient and temporal depth learning for face anti-spoofing. \textit{In Proceedings of the IEEE/CVF Conference on Computer Vision and Pattern Recognition} (pp. 5042-5051).

\bibitem{6}
Yu, Z., Zhao, C., Wang, Z., Qin, Y., Su, Z., Li, X., Zhao, G. (2020). Searching central difference convolutional networks for face anti-spoofing. \textit{In Proceedings of the IEEE/CVF Conference on Computer Vision and Pattern Recognition} (pp. 5295-5305).

\bibitem{7}
Muhammad, U., Yu, Z., Komulainen, J. (2022). Self-supervised 2D face presentation attack detection via temporal sequence sampling. \textit{Pattern Recognition Letters}, \textbf{156}, 15-22.

\bibitem{8}
Qin, Y., Yu, Z., Yan, L., Wang, Z., Zhao, C., Lei, Z. (2021). Meta-teacher for face anti-spoofing. \textit{IEEE transactions on pattern analysis and machine intelligence}.


\bibitem{9}
Zhi, Y., Tong, Z., Wang, L., Wu, G. (2021). Mgsampler: An explainable sampling strategy for video action recognition. \textit{ In Proceedings of the IEEE/CVF International Conference on Computer Vision} (pp. 1513-1522).


\bibitem{10}
Basha, S. H., Pulabaigari, V., Mukherjee, S. (2022). An information-rich sampling technique over spatio-temporal CNN for classification of human actions in videos. \textit{Multimedia Tools and Applications}, 1-19.

\bibitem{11}
Zhang, Z., Yan, J., Liu, S., Lei, Z., Yi, D., Li, S. Z. (2012, March). A face antispoofing database with diverse attacks. \textit{In 2012 5th IAPR international conference on Biometrics (ICB)} (pp. 26-31). IEEE.


\bibitem{12}
Chingovska, I., Anjos, A., Marcel, S. (2012, September). On the effectiveness of local binary patterns in face anti-spoofing. \textit{In 2012 BIOSIG-proceedings of the international conference of biometrics special interest group (BIOSIG)} (pp. 1-7). IEEE.

\bibitem{13}
Wen, D., Han, H., Jain, A. K. (2015). Face spoof detection with image distortion analysis. \textit{IEEE Transactions on Information Forensics and Security}, 10(4), 746-761.

\bibitem{14}
He, K., Zhang, X., Ren, S., Sun, J. (2016). Deep residual learning for image recognition. \textit{In Proceedings of the IEEE conference on computer vision and pattern recognition} (pp. 770-778).

\bibitem{15}
Schuster, M., Paliwal, K. K. (1997). Bidirectional recurrent neural networks. \textit{IEEE transactions on Signal Processing}, 45(11), 2673-2681.

\bibitem{16}
Prechelt, L. (1998). Early stopping-but when?. \textit{In Neural Networks: Tricks of the trade} (pp. 55-69). Springer, Berlin, Heidelberg.

\bibitem{17}
He, K., Zhang, X., Ren, S., Sun, J. (2015). Delving deep into rectifiers: Surpassing human-level performance on imagenet classification. \textit{In Proceedings of the IEEE international conference on computer vision} (pp. 1026-1034).

\bibitem{18}
Boulkenafet, Z., Komulainen, J., Hadid, A. (2015, September). Face anti-spoofing based on color texture analysis. \textit{In 2015 IEEE international conference on image processing (ICIP)} (pp. 2636-2640). IEEE.

\bibitem{19}
Atoum, Y., Liu, Y., Jourabloo, A., Liu, X. (2017, October). Face anti-spoofing using patch and depth-based CNNs. \textit{In 2017 IEEE International Joint Conference on Biometrics (IJCB)} (pp. 319-328). IEEE.

\bibitem{20}
Muhammad, U., Hadid, A. (2019, June). Face anti-spoofing using hybrid residual learning framework. \textit{In 2019 International Conference on Biometrics (ICB)} (pp. 1-7). IEEE.

\bibitem{21}
Muhammad, U., Holmberg, T., de Melo, W. C., Hadid, A. (2019, September). Face Anti-Spoofing via Sample Learning Based Recurrent Neural Network (RNN). \textit{In BMVC} (p. 113).

\bibitem{22}
Song, X., Zhao, X., Fang, L., Lin, T. (2019). Discriminative representation combinations for accurate face spoofing detection. \textit{Pattern Recognition}, 85, 220-231.

\bibitem{23}
Tu, X., Ma, Z., Zhao, J., Du, G., Xie, M., Feng, J. (2020). Learning generalizable and identity discriminative representations for face anti-spoofing. \textit{ACM Transactions on Intelligent Systems and Technology (TIST)}, 11(5), 1-19.


\bibitem{24}
Quan, R., Wu, Y., Yu, X., Yang, Y. (2021). Progressive transfer learning for face anti-spoofing. \textit{IEEE Transactions on Image Processing}, 30, 3946-3955.

\bibitem{25}
Liu, Y., Jourabloo, A., Liu, X. (2018). Learning deep models for face anti-spoofing: Binary or auxiliary supervision. \textit{In Proceedings of the IEEE conference on computer vision and pattern recognition} (pp. 389-398).

\bibitem{26}
Jourabloo, A., Liu, Y., Liu, X. (2018). Face de-spoofing: Anti-spoofing via noise modeling. \textit{In Proceedings of the European conference on computer vision (ECCV)} (pp. 290-306).

\bibitem{27}
Yang, X., Luo, W., Bao, L., Gao, Y., Gong, D., Zheng, S., . Liu, W. (2019). Face anti-spoofing: Model matters, so does data. \textit{In Proceedings of the IEEE/CVF Conference on Computer Vision and Pattern Recognition} (pp. 3507-3516).

\bibitem{28}
Simonyan, K., Zisserman, A. (2014). Very deep convolutional networks for large-scale image recognition. \textit{arXiv preprint arXiv}:1409.1556.


\bibitem{29}
Zeiler, M. D., Fergus, R. (2014, September). Visualizing and understanding convolutional networks. \textit{In European conference on computer vision (pp. 818-833)}. Springer, Cham.

\bibitem{30}
Ribeiro, M. T., Singh, S., Guestrin, C. (2016, August). " Why should i trust you?" Explaining the predictions of any classifier. \textit{In Proceedings of the 22nd ACM SIGKDD international conference on knowledge discovery and data mining} (pp. 1135-1144).



\end{thebibliography}

\end{document}